\documentclass[letterpaper, 10 pt, journal,twoside]{IEEEtran}
\IEEEoverridecommandlockouts

\usepackage{graphicx} \graphicspath{ {figures/} }
\usepackage{amsmath,amssymb,mathabx,amsbsy,mathtools,etoolbox}
\usepackage[utf8]{inputenc} % allow utf-8 input
\usepackage[T1]{fontenc}    % use 8-bit T1 fonts

\usepackage{algorithmic}
\usepackage[ruled]{algorithm2e}

\usepackage{acronym}
\usepackage{enumitem}
\usepackage{balance}
\usepackage{xspace}
\usepackage{setspace}
\usepackage[skip=3pt,font=small]{subcaption}
\usepackage[skip=3pt,font=small]{caption}
\usepackage[dvipsnames,svgnames,x11names]{xcolor}
\usepackage[capitalise,nameinlink]{cleveref}
\usepackage{booktabs,tabularx,colortbl,multirow,multicol,array,makecell}
\usepackage{pifont}
\usepackage{cite}
\usepackage{nicematrix}
\usepackage{bm}
\usepackage{dblfloatfix}
\makeatletter
\DeclareRobustCommand\onedot{\futurelet\@let@token\@onedot}
\def\@onedot{\ifx\@let@token.\else.\null\fi\xspace}

\def\ie{\emph{i.e}\onedot}

\def\etal{\emph{et al}\onedot}
\makeatother

\makeatletter
\def\BState{\State\hskip-\ALG@thistlm}
\makeatother

\makeatletter
\renewcommand{\paragraph}{%
  \@startsection{paragraph}{4}%
  {\z@}{0ex \@plus 0ex \@minus 0ex}{-1em}%
  {\hskip\parindent\normalfont\normalsize\bfseries}%
}
\makeatother

\crefname{algorithm}{Alg.}{Algs.}
\Crefname{algocf}{Algorithm}{Algorithms}
\crefname{section}{Sec.}{Secs.}
\Crefname{section}{Section}{Sections}
\crefname{table}{Tab.}{Tabs.}
\Crefname{table}{Table}{Tables}
\crefname{figure}{Fig.}{Fig.}
\Crefname{figure}{Figure}{Figure}
\definecolor{gblue}{HTML}{4285F4}
\definecolor{gred}{HTML}{DB4437}
\definecolor{ggreen}{HTML}{0F9D58}

\definecolor{mygray}{gray}{.92}

\acrodef{qp}[QP]{Quadratic Programming}
\acrodef{dof}[DoF]{Degree of Freedom}
\acrodef{ros}[ROS]{Robot Operating System}
\acrodef{ik}[IK]{Inverse Kinematics}
\acrodef{dof}[DoF]{Degree of Freedom}
\acrodef{com}[CoM]{Center of Mass}
\acrodef{cdm}[CDM]{Centroidal Dynamics Model}
\acrodef{cmm}[CMM]{Centroidal Momentum Matrix}
\acrodef{mpc}[MPC]{Model Predictive Control}
\acrodef{cop}[CoP]{Center of Pressure}
\acrodef{ocp}[OCP]{Optimal Control Problem}
\acrodef{micp}[MICP]{Mixed-Integer Convex Programming}
\acrodef{mip}[MIP]{Mixed-Integer Programming}
\acrodef{sdf}[SDF]{Signed Distance Field}
\acrodef{mlt}[MLT]{multi-location task}

\title{Path and Motion Optimization for Efficient Multi-Location Inspection with Humanoid Robots}

\author{Jiayang Wu$^*$, Jiongye Li$^*$, Shibowen Zhang, Zhicheng He, Zaijin Wang, Xiaokun Leng,\\ Hangxin Liu,~\IEEEmembership{Member,~IEEE}, Jingwen Zhang,~\IEEEmembership{Member,~IEEE}, Jiayi Wang,\\ Song-Chun~Zhu,~\IEEEmembership{Fellow,~IEEE}, Yao Su,~\IEEEmembership{Member,~IEEE}% <-this % stops a space
\thanks{This work was supported in part by the National Natural Science Foundation of China (No. 62403064, 62403063), and Shenzhen Special Fund for Future Industrial Development (No. KJZD20230923114222045). \textit{(Jiayang Wu and Jiongye Li contributed equally to this work.)}  \textit{(Corresponding author: Yao Su).}}
\thanks{Jiayang Wu, Jiongye Li, Shibowen Zhang, Song-Chun Zhu, Hangxin Liu, Jingwen Zhang, Jiayi Wang, and Yao Su are with State Key Laboratory of General Artificial Intelligence, Beijing Institute for General Artificial Intelligence (BIGAI), Beijing 100080, China (e-mails: wujiayang@bigai.ai; lijiongye@bigai.ai; zhangshibowen@bigai.ai; sczhu@bigai.ai; hxliu@bigai.ai; zhangjingwen@bigai.ai; wangjiayi@bigai.ai; suyao@bigai.ai).}
\thanks{Jiayang Wu, Jiongye Li, and Song-Chun Zhu are also with Department of Automation, Tsinghua University, Beijing 100084, China.}
\thanks{Shibowen Zhang is also with Department of Automation, University of Science and Technology of China, Hefei 230022, China.}
\thanks{Zhicheng He and Xiaokun Leng are with Department of Computer Science, Harbin Institute of Technology, Harbin 150001, China (e-mails:lengxiaokun@hit.edu.cn;  hezhicheng@hit.edu.cn).}
\thanks{Song-Chun Zhu is also with Institute for Artificial Intelligence and School of Artificial Intelligence, Peking University, Beijing 100871.}}

\begin{document}

\maketitle

\begin{abstract}
This paper proposes a novel framework for humanoid robots to execute inspection tasks with high efficiency and millimeter-level precision. The approach combines hierarchical planning, time-optimal standing position generation, and integrated \ac{mpc} to achieve high speed and precision. A hierarchical planning strategy, leveraging \ac{ik} and \ac{mip}, reduces computational complexity by decoupling the high-dimensional planning problem. A novel MIP formulation optimizes standing position selection and trajectory length, minimizing task completion time. Furthermore, an MPC system with simplified kinematics and single-step position correction ensures millimeter-level end-effector tracking accuracy. Validated through simulations and experiments on the Kuavo 4Pro humanoid platform, the framework demonstrates low time cost and a high success rate in multi-location tasks, enabling efficient and precise execution of complex industrial operations. 
\end{abstract}

\section{Introduction}

\setstretch{0.98}

\begin{figure}[ht!]
    \centering
    \includegraphics[width=\linewidth]{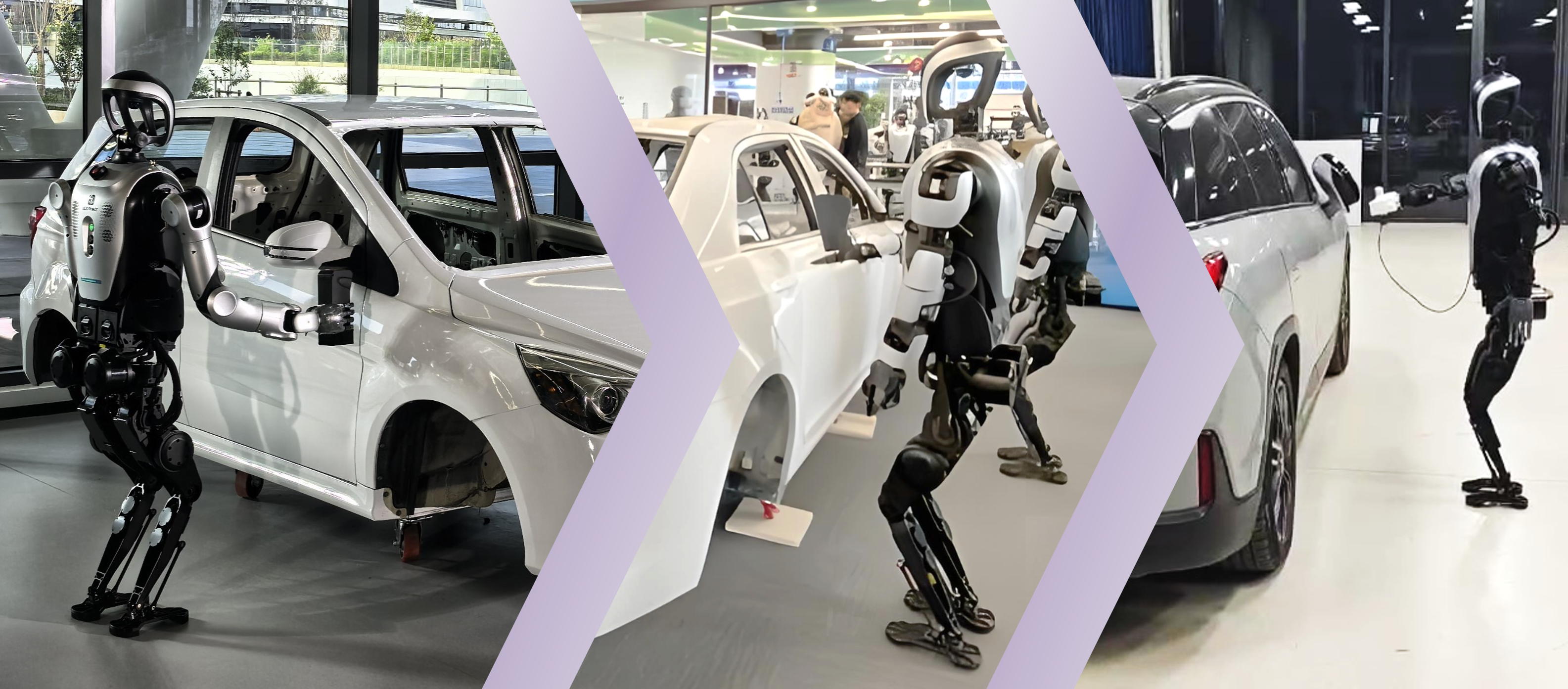}
     \caption{\textbf{Performing an inspection task with a humanoid robot.} The Dimension Technical Specifications (DTS) task in automotive quality check is utilized as an example, where the humanoid robot needs to hold a high-accuracy 3D scanner to reach multiple targets efficiently to measure the gap between different installation parts.}
     \vspace{-10pt}
    \label{fig:teaser}
\end{figure}

\IEEEPARstart{I}{n} large-scale manufacturing, such as automotive assembly, inspection is a critical stage for verifying product quality and ensuring adherence to design specifications. 
This process typically requires access to multiple locations to perform actions such as probing, soldering, and screw tightening. 
Due to the repetitive and labor-intensive nature of these duties, humanoid robots are emerging as a promising solution to take over these tasks from human workers~\cite{kheddar2019humanoid,tong2024advancements}. For instance, \cref{fig:teaser} illustrates a scenario in automotive quality assurance, where a humanoid robot repeatedly repositions itself to allow its arm to perform precise gap and flush measurements between adjacent body panels on the vehicle's exterior, thereby ensuring compliance with strict dimensional specifications.

On the other hand, when deploying humanoid robots for inspection tasks, an important issue is to find a path to guide the robot to complete the inspection process. 
This involves selecting both the specific standing positions to reach prescribed inspection targets and the sequence in which these positions are traversed.
While heuristic approaches---such as visiting inspection targets in an arbitrary order---offer simplicity, they often generate suboptimal paths.
Such paths may introduce unnecessary stops for the cases where multiple nearby inspection targets could be reached from a shared standing position, leading to excessive walk-to-stop transitions that incurs significant time penalties. Additionally, suboptimal paths may contain redundant loops, increasing overall travel distance. 
Following those suboptimal paths often leads to extended task execution time and reduced inspection efficiency, which is a critical concern in rapid manufacturing, particularly for long-horizon operations~\cite{tazaki2020survey}.

To enbale efficient multi-location inspection with humanoid robots, we decompose the problem into three hierarchical stages that optimize the traversing path of the robot. (1) \textbf{Feasible-region generation}: we first treat the humanoid’s body as a 2-\ac{dof} virtual base on the horizontal plane and perform \acf{ik} sampling for each prescribed inspection pose~\cite{jiao2021consolidating,su2023sequential,li2024dynamic}. This yields a dense set of feasible base positions whose union often forms a nonconvex region for each inspection point. (2) \textbf{Tolerance-circle abstraction}: for easing the computational burden imposed by nonconvexity, we introduce the concept of tolerance circle, the largest possible circle inside the aforementioned feasible regions. In areas where multiple feasible regions overlap, we likewise extract maximal circles that lie entirely within the intersection, leading to a single stance that can service multiple inspection points which can potentially reduces the number of walk-to-stop transitions. (3) \textbf{Optimal sequence planning}: a \acf{mip} problem is formulated to select an optimal subset of these tolerance circles as the robot’s standing positions while minimizing the number of required positions and the total base translation distance~\cite{ding2018single,deits2014footstep}.

Following the path planning stage, we employ a \ac{cdm}-based \acf{mpc} controller~\cite{he2024cdmpc,zhang2023design} with hybrid commands to generate whole-body robot motions that track the optimized inspection path. Specifically, body-velocity commands are used while the robot walks towards each planned standing position, which enables fast and smooth locomotion during most time of the task execution. However, when the robot gets close to the inspection targets, using velocity guided controller can result in frequent small-step adjustments, which introduce significant time penalties. To address this issue, we switch to a position-based interface that commands the robot to generate a precise swing leg trajectory, allowing the robot to step directly into the tolerance circle.

To evaluate the performance of our approach, we conduct simulation studies on an automotive quality inspection task using the Kuavo 4Pro humanoid robot. Compared to a naïve stop-and-go strategy where the robot sequentially walks towards designated positions and stops in front of each inspection target, our method reduces task completion time by 30\%. Moreover, computation time grows only linearly with the number of inspection targets, ensuring rapid planning even as task complexity increases. Finally, we validate the framework on a real-world vehicle assembly-gap inspection, demonstrating efficient multi-location inspection with millimeter-level
end-effector tracking accuracy. Our contributions in this work can be summarized as follows:

\begin{itemize}
    \item We introduce a hierarchical path optimization approach for improved task efficiency. 
    Our method combines \ac{ik} sampling and \ac{mip} to identify standing positions that can reach multiple targets without repositioning, and optimize the traversing path between them. 
    
    \item With decomposed planning and control modules, we propose a complete framework for efficient multi-location inspection with humanoid robots. 30\% reduction in task completion time is achieved compared to the baseline approach.
    
    %We conduct simulation studies on an automotive quality inspection task. Our method achieves a 30\% reduction in task completion time, comparing to a stop-and-go approach where the robot sequentially visits the inspection targets at a fixed distance. 
    
    \item We validate our approach in a real-world vehicle gap inspection task using the Kuavo 4Pro robot, demonstrating efficient multi-location inspection under practical conditions. 
                
\end{itemize}

\subsection{Related Work}
Early humanoid motion planners for potential industrial applications have decoupled locomotion from manipulation and relied on task‐level hierarchies or graph searches. Bouyarmane \etal introduced a best‐first, multi‐contact search that optimizes static equilibrium at each stance but faced scalability challenges in high-DoF systems~\cite{bouyarmane2012humanoid}. Murooka \etal combined reachability maps with graph search to sequence footsteps and arm reaches for tasks such as object retrieval, improving sequential task planning yet still treating walking and reaching independently~\cite{murooka2021humanoid}. GCS*, a graph-search variant on convex sets, enabled mixed discrete–continuous planning for contact sequencing but its implicit continuous search incurs heavy computation in non‐convex industrial workspaces~\cite{chia2024gcs}. Foundational decoupling strategies date back to Lozano‐Perez’s C‐space formulation~\cite{lozano1984approach}, which separated base and manipulator planning to reduce dimensionality, and were later refined by Kanoun \etal formulated footstep placement as an inverse‐kinematics optimization—yielding faster, but purely lower‐body, footstep generators. Yoshida \etal finally extended these ideas to whole‐body motion by stitching kinematic paths into coherent locomotion–manipulation trajectories~\cite{yoshida2010planning}, yet still without explicit handling of cumulative walking errors or task‐level end‐effector feasibility~\cite{gu2025humanoid}.

More recent work has extended optimal control problems (OCPs) to further improve locomotion robustness and versatility, but most have remained focused on lower-body motion alone and disregarded the kinematic and task constraints imposed by upper-body manipulation. Deits \etal formulated footstep planning on uneven terrain as a mixed‐integer convex program (MICP), achieving stable yet purely locomotion‐centric solutions~\cite{deits2014footstep}. Marcucci \etal extended this approach with convex relaxations tailored for cluttered, obstacle‐rich environments~\cite{marcucci2023motion}. Wang \etal introduced NAS, an N-step algorithm that exhaustively searches for feasible footstep sequences, guaranteeing completeness but at the expense of any end‐effector task integration~\cite{wang2024n}. In contrast, our framework co‐designs footstep planning and end‐effector control—simultaneously accounting for walking uncertainties and upper‐body task constraints—to generate standing positions and traversal paths that both minimize stop‐and‐go cycles and ensure manipulation feasibility for deployable industrial humanoid applications.

% \subsection{Overview}
% We organize the rest of the paper as follows. \cref{sec:problem} introduces the problem definition and preliminary of this work. \cref{sec:waypoints} proposes the standing position sequence planning framework. \cref{sec:control} introduces the \ac{mpc}-based controller design for the humanoid platform. \cref{sec:exp} presents the simulation a nd experiment results with comprehensive evaluations. Finally, we conclude the paperin \cref{sec:conclusion}.

\section{Preliminary \& Problem Definition}
\label{sec:problem}
\subsection{Dynamics Model}
The \ac{cdm} provides a compact representation of humanoid robot dynamics while preserving whole-body momentum characteristics~\cite{he2024cdmpc,zhang2023design}. This model establishes the relationship between generalized coordinates and centroidal momentum through the \ac{cmm} as:
\begin{equation}
    \pmb{H}=
    \begin{bmatrix}
      M\pmb{\dot{r}}\\ \pmb{h} 
    \end{bmatrix}=
    \pmb{A}_q(\pmb{q}) \pmb{\dot{q}},
    \label{eq:cmcal}
\end{equation}
where $\pmb{H}\in\mathbb{R}^{6\times1}$ denotes the centroidal momentum vector comprising linear ($M\pmb{\dot{r}}$) and angular ($\pmb{h}\in\mathbb{R}^{3\times1}$) components. The system parameters include total mass $M$, \ac{com} position $\pmb{r}\in\mathbb{R}^{3\times1}$, and configuration-dependent \ac{cmm} $\pmb{A}_q$ with $\pmb{q}\in\mathbb{R}^{6+n_q}$ as the generalized position vector, $n_q$ is the number of joints. 

Furthermore, Newtonian mechanics governs the momentum rate equations as:
\begin{equation}
\begin{bmatrix}
M\pmb{\ddot{r}} \\\dot{\pmb{h}}
 \end{bmatrix}=
\begin{bmatrix}
M\pmb{g}+\sum_{i=1}^{n_c}\pmb{f}_{{c}_{i}} \\
\sum_{i=1}^{n_c}(\pmb{{c}}_{i}-\pmb{r})\times \pmb{f}_{{c}_{i}}+\pmb{\tau}_{{c}_i}
\end{bmatrix}
\label{eq:momentrate}
\end{equation}
where $M\pmb{g}$ represents gravitational force, $\{\pmb{f}_{{c}_i},\pmb{\tau}_{{c}_i}\}\in\mathbb{R}^{3\times1}$ denote contact wrenches, and $\pmb{c}_i\in\mathbb{R}^{3\times1}$ specifies contact locations for ${n_c}$ active contact points.

\subsection{Time-optimal Trajectory Optimization}
To accomplish the inspection tasks efficiently, we formulated a time-optimal whole-body trajectory optimization problem as follows: 
\begin{subequations}
\begin{align}
    \min_{\pmb{Q},T_{\text{task}}} &\quad  T_{\text{task}}\\
 \text{s.t.} \quad & \bigcup_{t=1}^{T_{\text{task}}} \pmb{\mathcal{F}}_{\text{kin}}(\pmb{q}_{[t]}) \supseteq \pmb{\mathcal{G}} \label{eq:pro1}\\ 
 & \pmb{Q} = {\pmb{q}_{[1:T_{\text{task}}]}} \in \mathcal{C}_{\text{free}}\label{eq:pro3},\\
 & {\pmb{Q},\dot{\pmb{Q}},\ddot{\pmb{Q}}} \in \mathcal{D}_{\text{feas}},\label{eq:pro2}
\end{align}
\label{eq:timeoptimal_formulation}
\end{subequations}
where \( \pmb{\mathcal{G}} = \{\pmb{g}_{{\text{goal}},i}\}_{i=1}^{n}\in SE(3)\) represents target end-effector poses, $\pmb{Q}$ denotes the joint trajectory spanning $T_{\text{task}}$ timesteps, $\mathcal{C}_{\text{free}}$ defines the collision-free configuration space, and $\mathcal{D}_{\text{feas}}$ incorporates dynamic feasibility constraints. The forward kinematics operator $\pmb{\mathcal{F}}_{\text{kin}}(\cdot)$ maps joint configurations to end-effector poses.

Having this initial formulation, three fundamental challenges complicate the planning of inspection tasks: (i) \textbf{High-dimensional search space}: The 30+ \acp{dof} configuration space renders exhaustive search strategies computationally prohibitive; (ii) \textbf{Manipulation-mobility tradeoff}: Competing requirements for precise end-effector positioning versus efficient base locomotion create non-convex optimization landscapes; (iii) \textbf{Combinatorial sequencing}: The factorial growth of possible target visitation orders demands efficient permutation-space exploration.

% \subsection{Modeling Assumptions}
% Our framework incorporates the following physically-motivated assumptions: (i) The robot torso maintains quasi-static planar orientation during task execution (2D planar motion); (ii) Locomotion phase transitions are decoupled from manipulation tasks; (iii) Positional error in foot placement follows a zero-mean Gaussian distribution with $\sigma \leq 3cm$; (iv) Contact surfaces provide sufficient friction ($\mu \geq 0.5$) for stable locomotion.

\section{Standing Position Planning Pipeline}
\label{sec:waypoints}

\begin{figure}[t!]
    \centering
    \includegraphics[width=\linewidth]{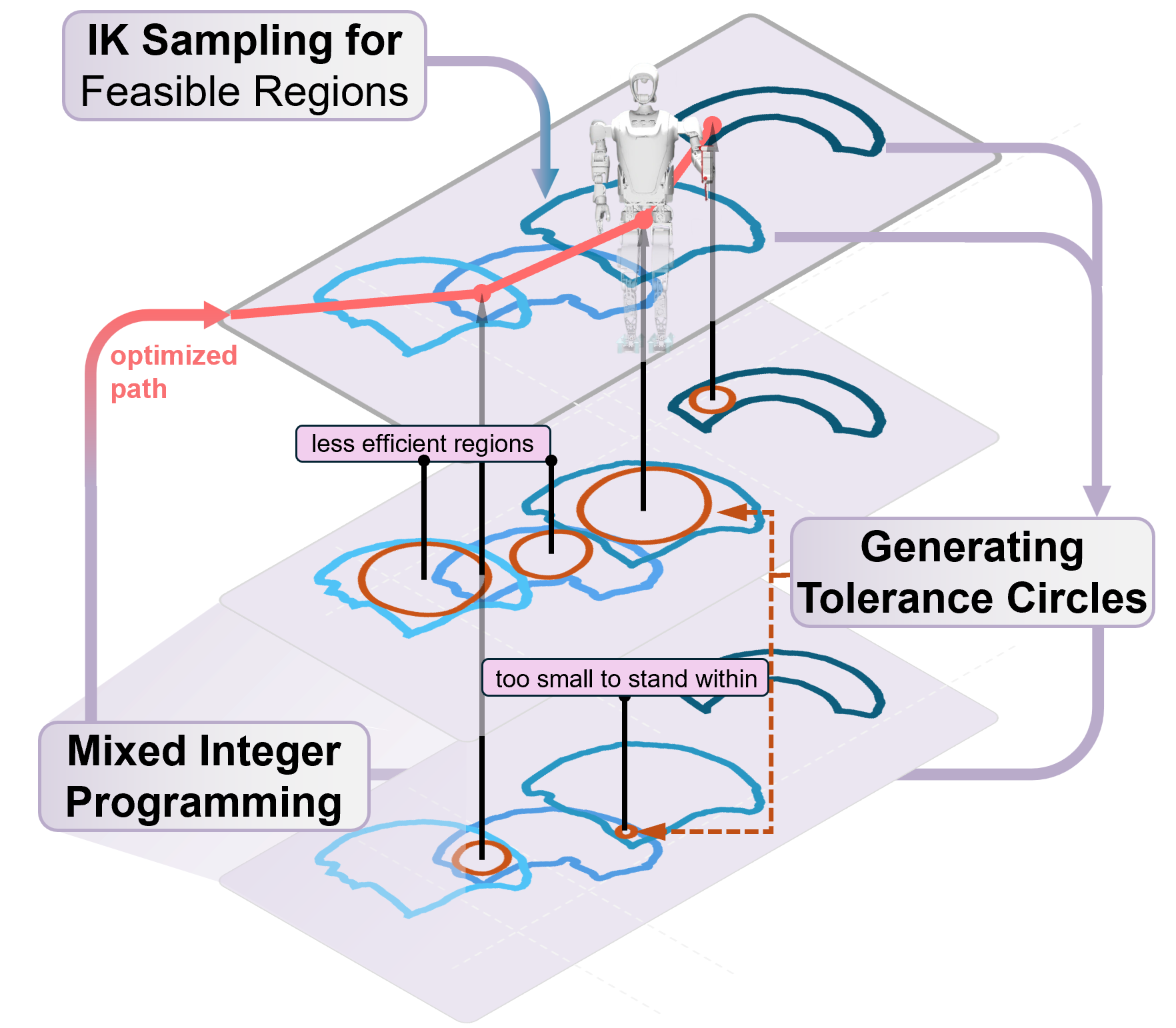}
    \caption{\textbf{The proposed MIP-based standing position planning pipeline.} Given a sequence of target end-effector poses, concave feasible base regions are computed via IK sampling. Within each region and their overlaps, the largest valid tolerance circles are generated to capture robust standing areas. A mixed-integer program then selects an optimal subset of circles and computes a walking path that minimizes both the number of standing positions and total travel distance, enabling efficient inspection.}
    \label{fig:waypoints}
\end{figure}

To address locomotion uncertainties and reduce stop-and-go cycles, we reformulate the original problem as an optimal stance sequence planning problem. Our three-stage solution pipeline comprises: (i) Configuration space exploration through IK sampling; (ii) Stability margin quantification via Voronoi-based tolerance regions; (iii) \ac{mip} for temporal sequence optimization. The whole planning pipeline is illustrated in  \cref{fig:waypoints}.

\subsection{Inverse Kinematics Formulation}
To simplify planning complexity, we abstract the kinematics of the robot as a 9-\ac{dof} virtual kinematic chain the horizontal (2-\acp{dof} for the base, 7-\ac{dof} for the arm)~\cite{li2024dynamic}. Then we compute the configuration space of each target point with the following \ac{ik} formulation to determine the feasible regions for task points.
\begin{subequations}
\begin{align}    
\min_{\pmb{s}_{xy},\pmb{q}^{a}} \; J_{\text{task}}(\pmb{s}_{xy}, \pmb{q}^{a}) +  \sum_{t=1}^{T} J_{\text{smooth}}(\dot{\pmb{q}}^{a}_{[t]})\\
  \pmb{q}^{a-} \leq \pmb{q}^{a}_{[t]} \leq \pmb{q}^{a+}, 
  \dot{\pmb{q}}^{a-} \leq \dot{\pmb{q}}^{a}_{[t]} \leq \dot{\pmb{q}}^{a+}, \forall t \in [1, T],\label{eq:cons3}\\
  \ddot{\pmb{q}}^{a-} \leq \ddot{\pmb{q}}^{a}_{[t]} \leq \ddot{\pmb{q}}^{a+}, \dddot{\pmb{q}}^{a-} \leq \dddot{\pmb{q}}^{a}_{[t]} \leq \dddot{\pmb{q}}^{a+}, \forall t \in [1, T],\label{eq:cons4}\\ 
  \pmb{q}^{a}_1 = \hat{\pmb{q}}^{a}_1, \quad \pmb{q}^{a}_T = \hat{\pmb{q}}^{a}_T,\label{eq:cons5}\\
   J_r(\pmb{\mathcal{F}}_{\text{kin}}(\pmb{s}_{xy},\pmb{q}^{a}_{[t]})) \geq 0, \forall t \in [1, T],\label{eq:cons6}\\
  J_e(\pmb{\mathcal{F}}_{\text{kin}}(\pmb{s}_{xy},\pmb{q}^{a}_{[t]})) \geq 0,\forall t \in [1, T],\label{eq:cons7} 
\end{align}
\label{eq:ik}
\end{subequations}
where the objective function minimizes a combination of task-specific costs ($J_{\text{task}}$) and motion smoothness ($J_{\text{smooth}}$).  $\pmb{s}_{xy}$ represents the virtual base position, and $\pmb{q}^{a}$ represents the arm joint angles.  The constraints include: joint limits on position, velocity, acceleration, and jerk; initial and final joint configurations that match task requirements; and collision avoidance constraints ($J_r$ and $J_e$) based on the \ac{sdf}~\cite{oleynikova2016signed} to prevent self and environmental collisions.

For each target point, we generate $10^5$ valid IK solutions via solving a batched optimization of \cref{eq:ik} in parallel with GPU-accelerated CuRobo~\cite{sundaralingam2023curobo}. The feasible base positions undergo geometric processing to construct concave hulls, representing the feasible regions of the robot’s base. This process uses the Shapely library~\cite{Gillies_Shapely_2025} and employs multi-process computation to handle multiple target points parallelly. The resulting concave hulls capture the geometric boundaries of the robot’s feasible base positions. By following this procedure, we obtain a set of polygons $\{\mathcal{T}_i\}_{i=1}^{n}$ that define the feasible base regions for each target point. 

\subsection{Tolerance Circle Generation}
After obtaining the feasible polygon $\mathcal{T}_i$ for each task point, we identify overlapping regions $\mathcal{T}_{Xi}$, which correspond to areas suitable to reach multiple work points, and save the reachable target poses in a set $\mathcal{G}_{Xi}$. Let $N_i$ denote the size of $\mathcal{G}_{Xi}$, \ie the number of reachable poses for the $i^{th}$ base position. For Non-overlapping regions $\mathcal{T}_{Hi}$, it follows that $N_i = 1$. Since overlapping regions $\mathcal{T}_{Xi}$ significantly influence the minimization of configuration numbers, we prioritize these regions based on their corresponding $N_i$.

Considering the walking positional accuracy of humanoid robots, instead of sampling standing positions, we generate tolerance circles within the feasible regions that present permissible deviations from the exact location. Experimental results indicate that generating multiple sampling circles does not significantly improve solution quality but substantially increases computation time due to the heightened complexity of the planning problem. This is primarily because the feasible region is relatively small, which has minimal impact on the time cost, whereas the sampling time is largely dictated by the number of tolerance circles. Therefore, we use a Voronoi-based algorithm to find the largest inscribed circle (with its center $s_i$ and radius) within each feasible polygon, as shown in \cref{fig:waypoints}. This approach ensures that we select a single, optimal standing position within each region, balancing solution quality and computational efficiency. Then, a set of valid standing positions is acquired $\pmb{S} = \{\pmb{s}_1, \pmb{s}_2, ..., \pmb{s}_m\}$. 
\subsection{Optimal Sequence Planning}
To select the optimal standing position sequence from $\pmb{S}$, we formulate an \ac{mip} problem to minimize the time cost while ensuring all task constraints are satisfied. The objective function is designed as follows:
\begin{equation}
    \min \; \alpha \sum_{i} z_i + \frac{1}{v} \sum_{i,j} o_{i,j} \cdot d_{i,j}
\end{equation}
where $z_i\in\{0,1\}$ is indicator of whether node $i$ is visited, $o_{i,j}\in\{0,1\}$ is the edge activation indicator between nodes $i,j$, $d_{i,j}$ represent the distance between $\pmb{s}_i$ and $\pmb{s}_j$, $\alpha$ penalizes the number of selected standing positions, while $\frac{1}{v}$ weights the total travel distance, where ${v}$ is the robot’s average walking speed. This objective function minimizes the total cost, which consists of two components: the cost of standing position for robot to stop and perform actions at each position, represented by $\sum_{i} z_i$, and the cost of trajectory distance, represented by $\sum_{i,j} o_{i,j} \cdot d_{i,j}$. To ensure the generated standing position sequence satisfies all task requirements, we designed several constraints, which will be introduced next.
% The constraints ensure that all relevant standing positions are covered, flow conservation is maintained, and sub-tours are avoided. 
%This formulation guarantees the generation of an optimal set of standing positions that satisfy task requirements while minimizing the overall time cost.

\subsubsection{\textbf{Terminal Constraints}}
The start and end points are critical to the task's success,
\begin{equation}
    z_0 = z_{m+1} = 1,
\end{equation}
this constraint explicitly specifies that the start node $z_0$ and end node $z_{m+1}$ must both be visited. Additionally, to ensure the presence of at least one valid path in the trajectory, we define the following constraint:
\begin{equation}
    \pmb{W} \cdot \pmb{z} \geq 1,
\end{equation}
where $\pmb{W}$ is a binary matrix representing connectivity of nodes and $\pmb{z}=[z_1,z_2,\cdots,z_m]^\mathsf{T}$. This guarantees that a valid trajectory exists by enforcing the inclusion of at least one active path from the start point to the end point.

\subsubsection{\textbf{Flow conservation constraints}}
These constraints ensure that the robot enters and exits each standing position properly, maintaining the continuity of the trajectory. For the start node:
\begin{equation}
    \sum_{j} o_{0,j} = 1, \quad \sum_{j} o_{j,0} = 0,
\end{equation}
which ensures the robot initiates its trajectory exclusively from $z_0$. For the end node:
\begin{equation}
    \sum_{j} o_{m+1,j} = 0, \quad \sum_{j} o_{j,m+1} = 1,
\end{equation}
this ensures that the robot terminates its trajectory at the endpoint $z_n$. For intermediate nodes:
\begin{equation}
    \sum_{j} o_{i,j} = z_i, \quad \sum_{j} o_{j,i} = z_i, \quad \forall i \neq 0, m+1,
\end{equation}
this enforces flow conservation at all intermediate standing positions. The inflow to a node must equal the outflow, ensuring that the robot visits and leaves each standing position exactly once if it is part of the trajectory ($z_i = 1$).

\subsubsection{\textbf{Order and Sub-Tour Constraints}}
To maintain a valid sequence, we introduce node ordering constraints to ensure that the desired points are visited in the order required by the task requirements:
\begin{equation}
    \gamma_0 = 0, \quad \gamma_m = m+1,  \quad 1 \cdot z_i \leq \gamma_i \leq m \cdot z_i , \quad \forall i \leq m.
\end{equation}
Here, $\gamma_i$ is a continuous variable that represents the order of visiting node $i$. The constraints ensure: (1) The starting node is always assigned the first order ($\gamma_0 = 1$); (2) For any node $i$ that is part of the trajectory ($z_i = 1$), the order $\gamma_i\in[2,m]$; 
 (3) If a node $i$ is not visited ($z_i = 0$), then $\gamma_i$ is effectively constrained to zero.
 
% \subsubsection{\textbf{Prevent Sub-Tour Elimination}}
To avoid invalid solutions with sub-tours (cycles that do not include the start or end points), we impose the following constraint:
\begin{equation}
    \gamma_i - \gamma_j + (m+2) \cdot o_{i,j} \leq (m+1) , \forall 0 \leq i,j \leq m+1, i \neq j
\end{equation}
which enforces a consistent order of visiting nodes. If a path exists between nodes $i,j$ ($o_{i,j} = 1$), then the order $\gamma_i$ must be less than $\gamma_j$, maintaining the trajectory's continuity.
\subsubsection{\textbf{Total Flow Constraints}}
The total flow constraint ensures that the number of active paths in the solution matches the number of active nodes minus one (the starting node has no incoming path):
\begin{equation}
    \sum_{i,j} o_{i,j} = \sum_{i} z_i - 1.
\end{equation}
This constraint ties the flow variables $o_{i,j}$ to the node activation variables $z_i$. It ensures that the solution is consistent, with exactly one trajectory covering all the active nodes.

% \begin{figure*}
%     \centering
%     \includegraphics[width=1\linewidth,trim=0cm 5cm 0cm 0cm,clip]{figures/placeholder.png}
%     \caption{\textbf{The system diagram for the KUAVO 4Pro humanoid robot.} (a) humanoid robot's hardware configuration and communication diagram. (b) The control block diagram of the humanoid robot platform.}
%     \label{fig:system_config}
% \end{figure*}

\begin{figure*}
    \centering
    \includegraphics[width=1\linewidth,trim=0.8cm 6.8cm 9.3cm 4cm,clip]{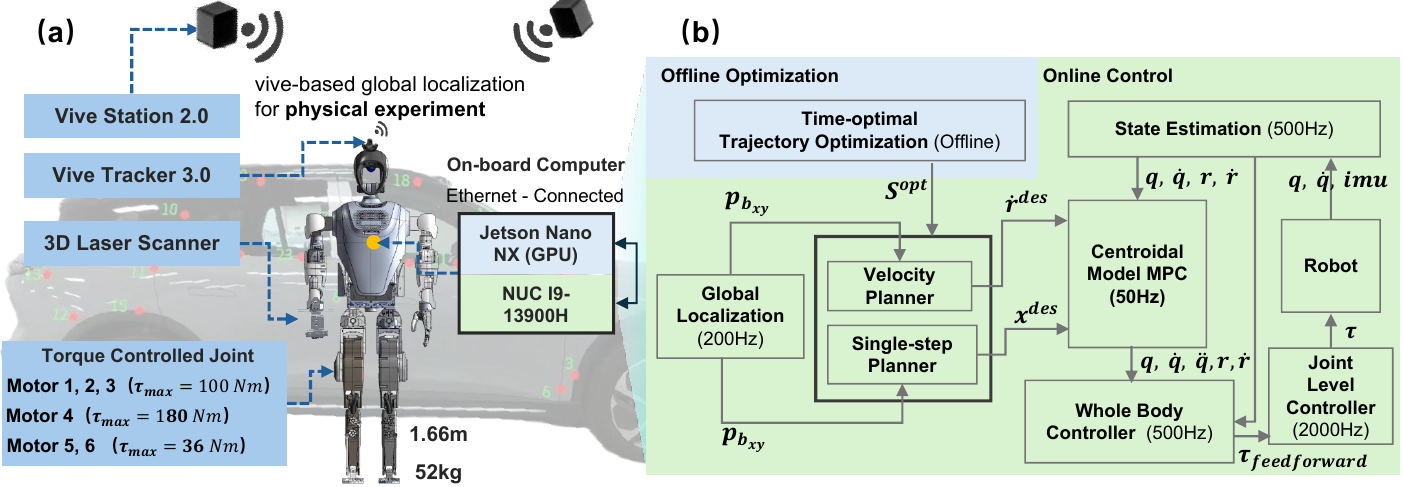}
    \caption{\textbf{The experimental setup and system diagram.} (a) The inspection task scenario in simulation and experiment, together with the hardware configuration of the bipedal humanoid robot KUAVO 4Pro. (b) The control block diagram of the inspection task with both velocity control and single-step control modes.}
    \label{fig:system_config}
\end{figure*}

\subsubsection{\textbf{Initial Guess}}
To guide the optimization process and prioritize nodes in overlapped regions, we formulate a warm-start strategy by constructing initial guesses for $z_i$ variables:
\begin{equation}
z_i^{init} =
\begin{cases}
1 & \text{if } \pmb{s}_i \in \mathcal{O}  \\
0 & \text{otherwise}
\end{cases}
\end{equation}
where $\mathcal{O}={\pmb{s}_i | \pmb{s}_i \in \bigcup \mathcal{T}_{Xi}}$ denotes the set of standing positions located in overlapped regions. This initialization is complemented with a priority weighting factor in the objective function:
\begin{equation}
\min  a_1 \sum_{i} z_i + \frac{1}{v} \sum_{i,j} o_{i,j} \cdot d_{i,j} - \lambda \sum_{i \in \mathcal{O}} z_i
\end{equation}
where $\lambda$ controls the bonus reward for selecting overlapped standing positions and is set to 10\% of $a_1$. 

 Finally, we obtain an ordered sequential path of standing positions,  denoted as $\pmb{S}^{\text{opt}} = \{\pmb{s}^{\text{opt}}_1, \pmb{s}^{\text{opt}}_2, \ldots, \pmb{s}^{\text{opt}}_{m'}\}, \pmb{s}^{\text{opt}}_i\in \pmb{S}$ in the 2D plane where $z_i = 1, m'= \sum_{i} z_i$, which is supplied to the model predictive controller.
 
\section{Model Predictive Control}
\label{sec:control}
Humanoid locomotion control faces a trade-off between precise step control and velocity-based control. The precise step control method performs well in high-precision scenarios by enforcing precise foot position and body trajectory but has limitations in terms of dynamic environment adaptability. Conversely, the velocity-based control method improves time efficiency through velocity commands, enabling rapid responses but accumulating positional drift and compromising foot step accuracy over time.

Therefore, this article proposes a hybrid nonlinear \ac{mpc} method that combines velocity-based control with precise step control. During steady-state locomotion, the system adopts a speed tracking-based control mode to ensure smooth motion. When the robot approaches the target position, the system automatically switches to a single-step control strategy (illustrated in \cref{fig:singlestep}), achieving terminal positioning accuracy control through precise foothold tracking and trajectory interpolation.

The controller solves a discrete-time \ac{ocp} over a receding horizon~\cite{wang2024online}. The objective function is designed as:
\begin{equation}
\resizebox{0.89\linewidth}{!}{
$\begin{aligned}
\min_{\pmb{u}_{[0:N-1]}} &  \sum_{k=0}^{N-1}\left(\left\| \pmb{x}_{[k]}-\pmb{x}_{[k]}^{des}\right\| _{\pmb{Q}}^2+\left\|\pmb{u}_{[k]}-\pmb{u}_{[k]}^{des}\right\|_{\pmb{R}}^2\right) +\left\|\pmb{x}_{[N]} - \pmb{x}_{[N]}^{\text{des}}\right\|_{\pmb{P}}^2
% \text{s.t.} \quad&\quad\quad\quad\quad\quad\quad  \pmb{x}_{[k+1]} = f(\pmb{x}_{[k]}, \pmb{u}_{[k]}) \\
% &\quad\quad\quad\quad\quad\quad \pmb{g}_{\text{kin}}(\pmb{x}_{[k]}, \pmb{u}_{[k]}) \leq 0 \\
% &\quad\quad\quad\quad\quad\quad \pmb{h}_{\text{dyn}}(\pmb{x}_{[k]}, \pmb{u}_{[k]}) = 0
\end{aligned}$}
\end{equation}
where $N$ is the total number of timesteps, $\|^\cdot\|_{[\cdot]}^2$ denotes the weighted squared $L_{2}$ norm, with $\pmb{Q},\pmb{R},\pmb{P}$ are weighting matrices. The reduced-order state vector $\pmb{x}$ is defined as:
\begin{equation}
\pmb{x}=\left[\pmb{\dot{r}}^\mathsf{T},\pmb{h}^\mathsf{T},\pmb{q}^\mathsf{T}\right]^\mathsf{T}\in\mathbb{R}^{12+{n_q}}.
\end{equation}
Control inputs $\pmb{u}$ include contact wrench at each contact point and joint velocities:
\begin{equation}
\pmb{u}=\left[\pmb{f}_{c_1}^\mathsf{T},\ldots,\pmb{f}_{c_{n_{c}}}^\mathsf{T},\pmb{\tau}_{c_1}^\mathsf{T},\ldots,\pmb{\tau}_{c_{n_c}}^\mathsf{T},\pmb{v}_j^\mathsf{T}\right]^\mathsf{T}\in\mathbb{R}^{6{n_c}+{n_q}}.
\end{equation}
\subsection{Constraint Architecture}
To ensure the feasibility and performance of the control system in actual operation, the \ac{mpc} formulation enforces the following constraints:

\textbf{The kinematic constraints} include:
\begin{subequations}
\begin{align}
    \pmb{H}_{[k]}&=\pmb{A}_q(\pmb{q}_{[k]}) \pmb{\dot{q}}_{[k]},\label{eq:planning_con1}\\
    \pmb{r}_{[k]}&=\pmb{\mathcal{F}} _r(\pmb{q}_{[k]}),\label{eq:planning_con2}\\
\pmb{c}_{{i}{[k]}}&=\pmb{\mathcal{F}}_{{c}_{i}}(\pmb{q}_{[k]}),\quad\quad\quad\quad\quad\,\, \forall i\in\{1,\ldots,n_{c}\},\label{eq:planning_con3}\\
\pmb{\dot{q}}_{j[k]}&=\pmb{v}_{j[k]},
\label{eq:planning_con4}\\
\pmb{v}_{ee[k]}&=\pmb{0},\,\,\quad\quad\quad\quad\quad\quad\quad\quad\quad\quad\quad \mathrm{if~}\pmb{p}_{ee} \in \pmb{S},
\label{eq:planning_con5}\\
\pmb{v}_{ee_z[k]}&=\pmb{v}_{ee_z[k]}^{des},\pmb{p}_{ee_z[k]}=\pmb{p}_{ee_z[k]}^{des},\quad\quad\mathrm{if~}\pmb{p}_{ee} \notin \pmb{S},
\label{eq:planning_con6}
\end{align}
\label{eq:kine_cons}
\end{subequations}
where $\pmb{\mathcal{F}}_r(\cdot)$ and $\pmb{\mathcal{F}}_{c_{i}}(\cdot)$  are forward kinematics functions to compute $\pmb{r}$ and $\pmb{c}_{i}$, $\pmb{S}$ is the collection of contact points, $\pmb{p}_{ee}$ and $\pmb{v}_{ee}$ is the positon and velocity of the end effector. Of note, \cref{eq:planning_con6} ensures accurate tracking of the end effector in the z-direction. \cref{eq:planning_con5} means the foot of a stance leg should not separate or slip with the ground.

To track the position of the end effector more accurately, we add soft constraints in the x and y directions on top of constraining the z-direction of the end effector as follows:
\begin{subequations}
\begin{align}
\pmb{v}_{ee_x[k]}=\pmb{v}_{ee_x[k]}^{des},\quad &\pmb{p}_{ee_x[k]}=\pmb{p}_{ee_x[k]}^{des},\quad\quad\mathrm{if~}\pmb{p}_{ee} \notin \pmb{S},
\label{eq:planning_ee1}\\
\pmb{v}_{ee_y[k]}=\pmb{v}_{ee_y[k]}^{des},\quad &\pmb{p}_{ee_y[k]}=\pmb{p}_{ee_y[k]}^{des},\quad\quad\mathrm{if~}\pmb{p}_{ee} \notin \pmb{S}.
\label{eq:planning_ee3}
\end{align}
\end{subequations}

\textbf{The dynamics constraints} include:
\begin{subequations}
\begin{align}
   \pmb{\dot{h}}_{[k]}&=\sum_{i=1}^{n_c}
	(\pmb{c}_{{i}{[k]}}-\pmb{r}_{[k]}) \times \pmb{f}_{c_{{i}[k]}}+\pmb{\tau}_{c_{{i}{[k]}}},\label{eq:planning_con8}  \\
    \pmb{\ddot{r}}_{[k]}&=\frac{1}{M}{{\sum_{i=1}^{n_c}{\pmb{f}_{c_{{i}[k]}}}}}+\pmb{g},\label{eq:planning_con9}\\
     \pmb{f}_{c_{{i}[k]}}&=\pmb{\tau}_{c_{{i}[k]}}=\pmb{0},\quad\forall i\in\{1,\ldots,n_{c}\},\mathrm{if~}\pmb{c}_{i[k]} \notin \pmb{S},
     \label{eq:planning_con10}\\
     \mu\pmb{f}_{c_{{i}[k]},z}&-\sqrt{\left({\pmb{f}_{c_{{i}[k]},x}}\right)^{2}+\left({\pmb{f}_{c_{{i}[k]},y}}\right)^{2}}\geq{0},
     \label{eq:planning_con12}
\end{align}
\label{eq:dyn_cons}
\end{subequations}
where $\mu$ represents the coefficient of friction. \cref{eq:planning_con12} maintains contact force within the friction cone.  \cref{eq:planning_con10} guarantees that the force and torque of the robot swing foot are zero.

\subsection{Velocity Planner}

Since the input of \ac{mpc} is the desired \ac{com} velocity $\pmb{\dot{r}}_{[k]}^\text{des}$, we introduce a velocity planner to generate the reference velocity profile along the planned path. The velocity planner computes $\pmb{\dot{r}}_{[k]}^\text{des}$ at each control step based on the current position $\pmb{r}_{[k]}$ and the next target position $\pmb{s}^{\text{opt}}_{i+1}$ on the path.

Specifically, at each time step, the direction vector towards the next target is determined as
\begin{equation}
    \pmb{d}_{[k]} = \pmb{s}^{\text{opt}}_{i+1} - \pmb{r}_{[k]}.
\end{equation}
The desired velocity is then computed using a PID controller applied to the distance error, and subsequently clipped by the maximum allowable velocity in each direction:
\begin{subequations}\label{eq:vel-planner}
\begin{equation}
    \pmb{\dot{r}}_{[k]}^{\text{pid}} = K_p \pmb{e}_{[k]} + K_i \sum \pmb{e}_{[k]}dt + K_d \frac{d\pmb{e}_{[k]}}{dt},
\end{equation}
\begin{equation}
    \pmb{\dot{r}}_{[k]}^{\text{des}} = \mathrm{clip}\left(\pmb{\dot{r}}_{[k]}^{\text{pid}}, -\pmb{v}^{\text{max}}, \pmb{v}^{\text{max}}\right),
\end{equation}
\end{subequations}
where $\pmb{e}_{[k]} = \frac{\pmb{d}_{[k]}}{\|\pmb{d}_{[k]}\|} \cdot \min\left(\|\pmb{d}_{[k]}\|, v^{\text{max}} \Delta t \right)$ is the limited direction error, $K_p$, $K_i$, and $K_d$ are the PID gains, $\pmb{v}^{\text{max}} = [v_x^{\text{max}}, v_y^{\text{max}}]^\top$ is the vector of maximum allowable velocities in the $x$ and $y$ directions, and $\mathrm{clip}(\cdot, a, b)$ limits each component of the vector to $[a, b]$. The resulting $\pmb{\dot{r}}^{\text{des}}$ is supplied to the MPC as the reference center of mass velocity at each step.
% In the \ac{mpc} formulation, the desired state $\pmb{x}_{[k]}^{{des}}$ is provided by the user via a command, and the control input is derived by solving the corresponding optimization problem.
\subsection{Single-Step Planner}

\begin{figure}[t!]
    \centering
    \includegraphics[width=0.8\linewidth,trim=0cm 0.5cm 0cm 0cm,clip]{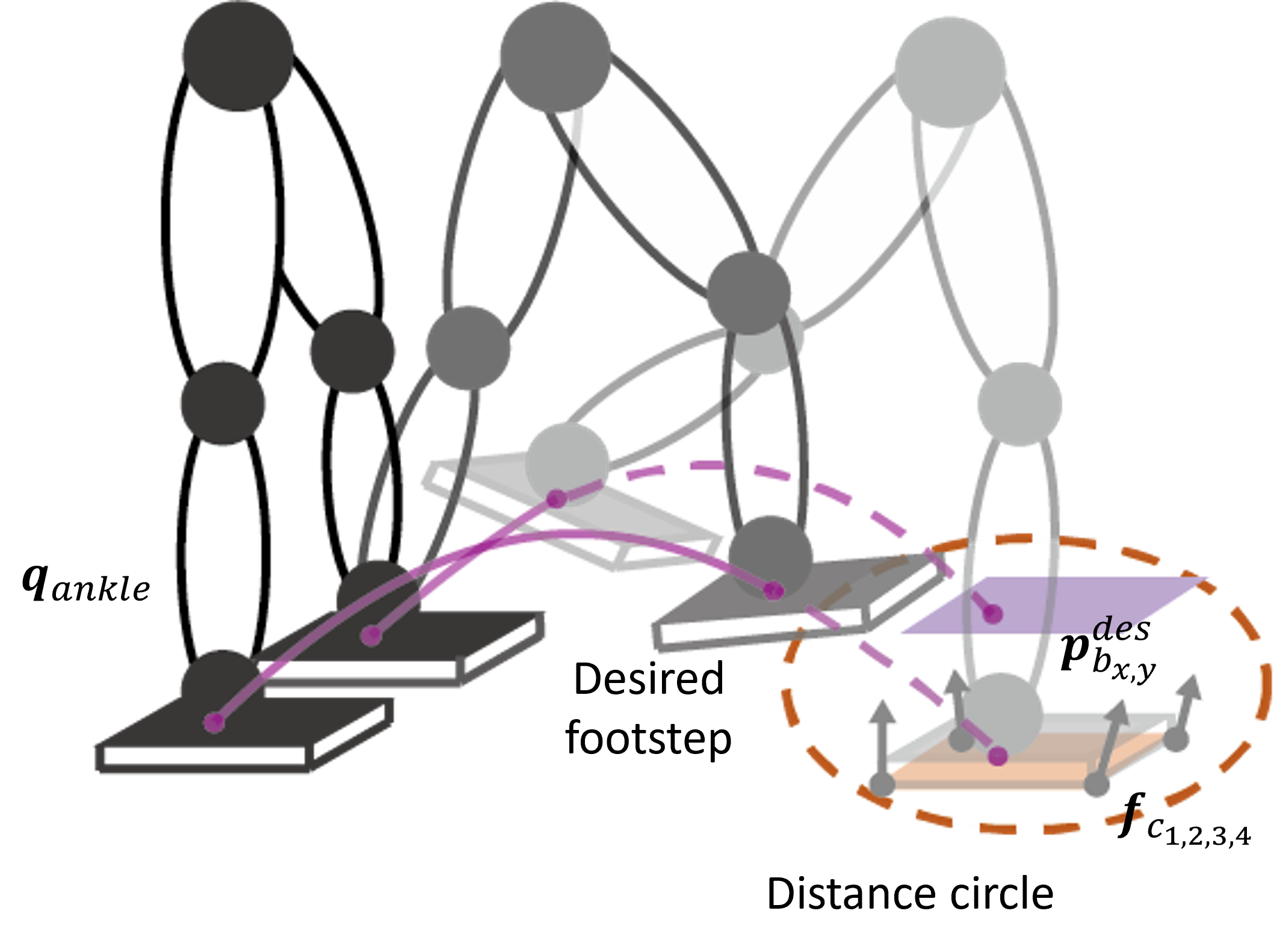}
    \caption{\textbf{Single-step control illustration.} If the robot's position is near the distance circle, single-step control will be used to generate new footsteps for more accurate tracking of target location $\pmb{p}_{b_{x,y}}^{des}$.}
    \label{fig:singlestep}
\end{figure}

As illustrated in \cref{fig:singlestep}, in single-step control, $\pmb{x}_{[k]}^{{des}}$ is defined as the desired body position $\pmb{p}_{b{[k]}}^{des}$ and yaw angle $\pmb{\theta}_{b_{y}{[k]}}^{{des}}$. When the relative distance and angle to the target position are within the threshold, the footstep is calculated. Then the trajectory of the target swing leg is obtained by performing cubic interpolation between the starting and target footsteps. This trajectory is used as soft constraints for the end effector's position $\pmb{p}_{ee}$ and velocity $\pmb{v}_{ee}$ in the MPC formulation, as shown in \cref{eq:planning_ee1} and \cref{eq:planning_ee3}. 

By embedding the hybrid control method into the \ac{mpc} framework, we achieved an effective balance between time efficiency and localization accuracy. This integrated solution enables robots to autonomously adapt to environmental changes and dynamically compensate for accumulated deviations during motion, significantly improving the overall control performance of the system.

\section{Evluation}
\label{sec:exp}

To rigorously validate the effectiveness of the standing position generation algorithm and multi-point whole-body planning and control framework proposed in this paper, we conducted algorithm validation experiments both in simulation and in the real world. 

\subsection{Humanoid Platform}

As shown in \cref{fig:system_config}a, the Kuavo 4Pro humanoid robot experimental platform, developed by Leju Robot, was utilized to validate the planning and control algorithms proposed in this study. This humanoid robot has a height of $1.75m$ and a mass of $50kg$. Each leg comprises six \acp{dof}, while each arm incorporates seven \acp{dof}. The robot operates on a Linux-based system augmented with a real-time kernel patch, ensuring efficient sensor data acquisition. The computational framework for control and learning algorithms are supported by two onboard computers: an Intel NUC (Core i9-13900H) and an NVIDIA Jetson AGX Orin (64GB). All joint actuators are operated in torque control mode, enabling precise and responsive motion control.

\subsection{Simulation Setups}

We conducted three simulation experiments. The first analyzed how the computation time of the proposed \ac{mip} planning method scales with the number of target poses and compare the corresponding execution time with a naive point‐to‐point planning method. The second evaluated the single‐step control accuracy of the robot as introduced in \cref{sec:control}. The third compared the robot’s actual performance in an automotive inspection task when using our proposed standing position‐generation method against the naïve point‐to‐point planning and control scheme.

In terms of implementation details, We used MuJoCo~\cite{todorov2012mujoco} as the simulation platform and applied an identified robot dynamic model. We have implemented the standing position sequence optimization algorithm utilizing Shapely~\cite{Gillies_Shapely_2025} to perform precise geometric computations and the commercial solver Gurobi~\cite{gurobi} to solve the resulting \ac{mip} formulation.

\subsection{Simulation Results}

\begin{figure}[ht!]
    \centering
    \includegraphics[width=1\linewidth]{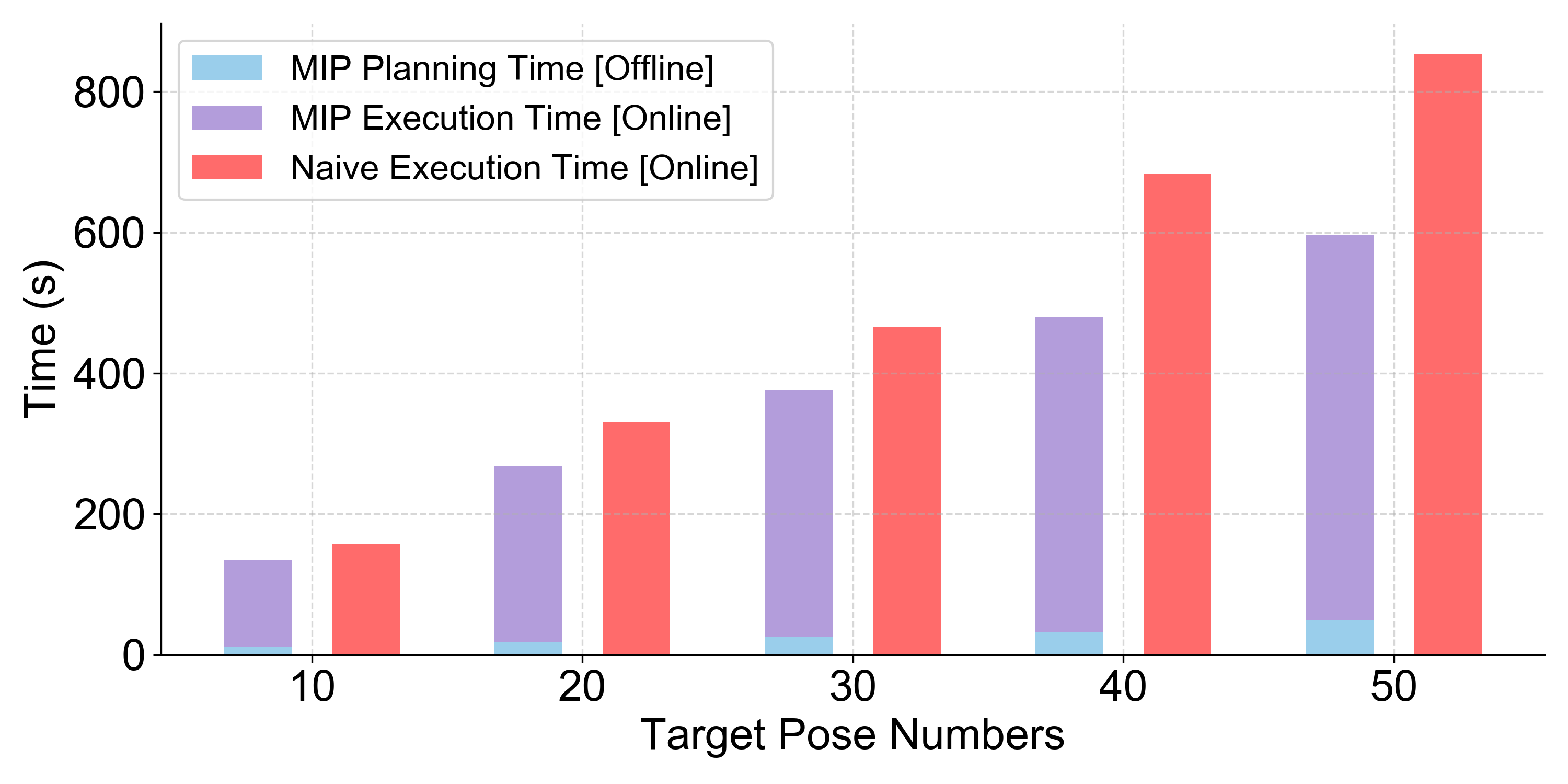}
    \caption{\textbf{The comparison of planning and execution time with the proposed MIP method and naive point‐to‐point planning method.}}
    \label{fig:time_plot}
\end{figure}

\subsubsection{Task Planning and Execution Time Analysis with Respect to Target Pose Quantity}

To evaluate the task planning and execution efficiency of our algorithm, we measured the planning and execution time of the proposed MIP method and a naive point‐to‐point baseline planning method, which computes and executes each target‐point trajectory in isolation, completing the plan for one standing position before initiating the next without any global optimization. with a different number of target poses. These target poses were generated by randomly sampling the robot's base position within a feasible region and assigning random joint configurations to its manipulator arm, resulting in a diverse set of end-effector goals.

As shown in \cref{fig:time_plot}, the proposed MIP method significantly reduces the online execution time compared to the naive point-to-point method, especially as the number of target poses increases. This improvement is attributed to the strategic selection of standing positions that minimize the overall travel distance and reduce unnecessary stops and starts. Specifically, the naive point-to-point strategy suffers from a linear growth in execution time, as it lacks global optimization and requires sequentially solving each subtask in real-time. In contrast, the MIP method leverages offline planning to compute globally optimized standing points, resulting in a much lower cumulative online execution cost. 

Furthermore, as shown in \cref{tab:computetime}, the overall computation time of the proposed \ac{mip}-based method increases linearly with the number of target poses, suggesting that the method scales well with task size. A breakdown of the computation time shows that the parallel \ac{ik} module dominates the total cost. The overlap calculation and tolerance circle generation modules exhibit moderate growth, while the MIP optimization remains lightweight for small-scale inputs but becomes more prominent as the task size increases. These results demonstrate that the proposed pipeline is computationally efficient even under randomized and high-variance input conditions.

\begin{table}[t!]
\centering
\small
\caption{\textbf{Mean com putation time of each stage within the proposed MIP-based standing position planning pipeline.} IK stands for the parallel inverse kinematic process, OC stands for the overlap calculation process and TCG represents the tolerance circle generation process.}
\begin{tabular}{c|ccccc}
\toprule
\multirow{2}{*}{Stage} & \multicolumn{5}{c}{Target Pose Numbers} \\
 & 10\rule{0pt}{2.5ex} & 20\rule{0pt}{2.5ex} & 30\rule{0pt}{2.5ex} & 40\rule{0pt}{2.5ex} & 50\rule{0pt}{2.5ex} \\
\midrule
IK   & $8.16s$  & $13.83s$ & $19.45s$ & $24.32s$ & $30.22s$ \\
OC   & $2.52s$  & $3.49s$  & $3.47s$  & $3.96s$  & $4.63s$ \\
TCG  & $0.62s$  & $1.09s$  & $1.57s$  & $2.06s$  & $2.51s$  \\
MIP  & $0.06s$  & $0.21s$  & $0.77s$  & $1.79s$  & $10.88s$ \\
\midrule
\textbf{Overall} & \textbf{$11.36s$} & \textbf{$18.62s$} & \textbf{$25.26s$} & \textbf{$32.13s$} & \textbf{$48.24s$} \\
\bottomrule
\end{tabular}
\label{tab:computetime}
\end{table}

\subsubsection{Ablation Study of Single‐Step Control Strategy}

We also performed an ablation study to investigate the impact of the proposed single-step control module on goal-directed locomotion accuracy.  For with and without the single-step control conditions, we conducted 30 repeated trials, where the humanoid robot was initialized from randomized starting positions and commanded to reach a fixed target at coordinates $(x = 1m, y = 0m)$ on the XY plane.

\begin{figure}[ht!]
    \centering
    \includegraphics[width=0.8\linewidth]{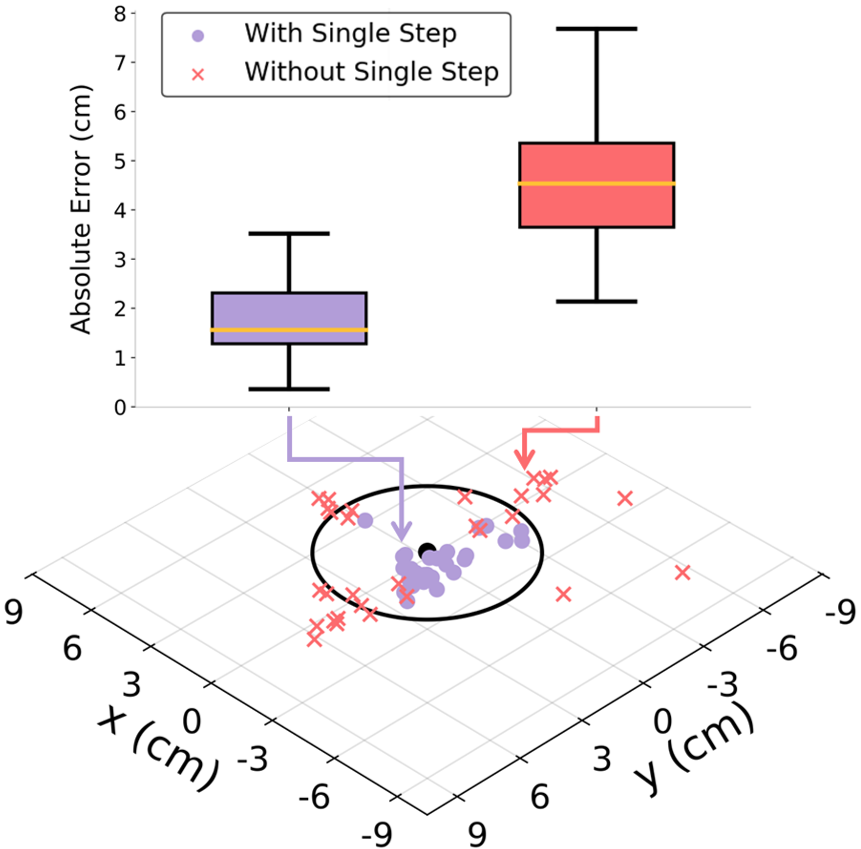}
    \caption{\textbf{Comparison of goal-reaching accuracy in random-start locomotion tasks with and without single-step control.} The lower scatter plot shows final positions in the XY plane, where trials with single-step control (purple) converge more tightly near the target than those without (red). The upper box plot confirms lower median error and variance with single-step control. The black circle in the scatter plot represents the minimum tolerance circle radius threshold for acceptable terminal accuracy.}
    \label{fig:singlestep_compare}
\end{figure}

As shown in \cref{fig:singlestep_compare}(a), the scatter plot illustrates the final positions reached by the robot across the 10 trials under both conditions. The blue dots, representing trials with Single Step Control, are tightly clustered around the target point $(x = 1m, y = 0m)$, indicating consistent and accurate convergence. In contrast, the red crosses, representing trials without single-step control, show significant spatial dispersion in the X and Y directions, reflecting decreased stability and precision in the absence of the proposed control strategy.

\cref{fig:singlestep_compare}(b) quantitatively compares the terminal XY-plane position errors. The results clearly demonstrate that single-step control reduces both the mean error and variance across the 30 repeated runs. In the absence of this control, the error distribution becomes broader with higher outliers exceeding $0.08m$, highlighting the importance of stepwise correction in trajectory tracking. 

\subsubsection{Automotive Inspection Task Validation}

\begin{figure}[t!]
    \centering
    \begin{subfigure}[b]{\linewidth}
        \includegraphics[width=\linewidth,trim=1cm 0cm 3cm 0cm,clip]{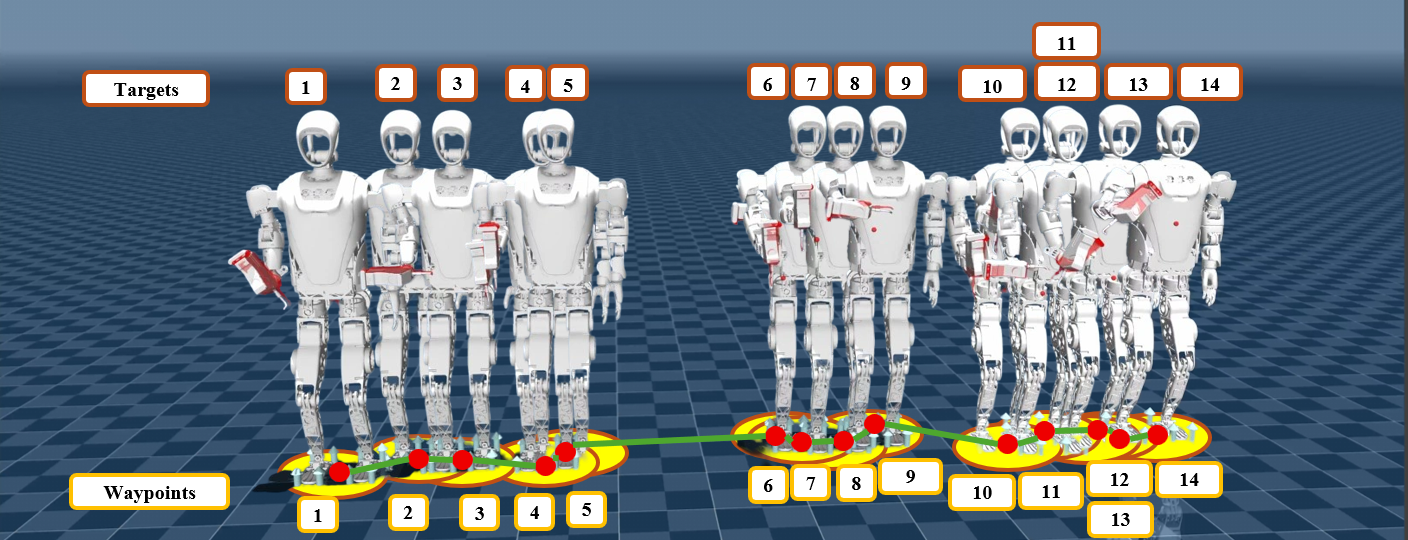}
        \caption{Naive point‐by‐point planning: 14 poses scanned in $265s$.}
        \label{fig:sim_exp_naive}
    \end{subfigure}\\
    \begin{subfigure}[b]{\linewidth}
        \includegraphics[width=\linewidth,trim=1cm 0cm 3cm 1cm,clip]{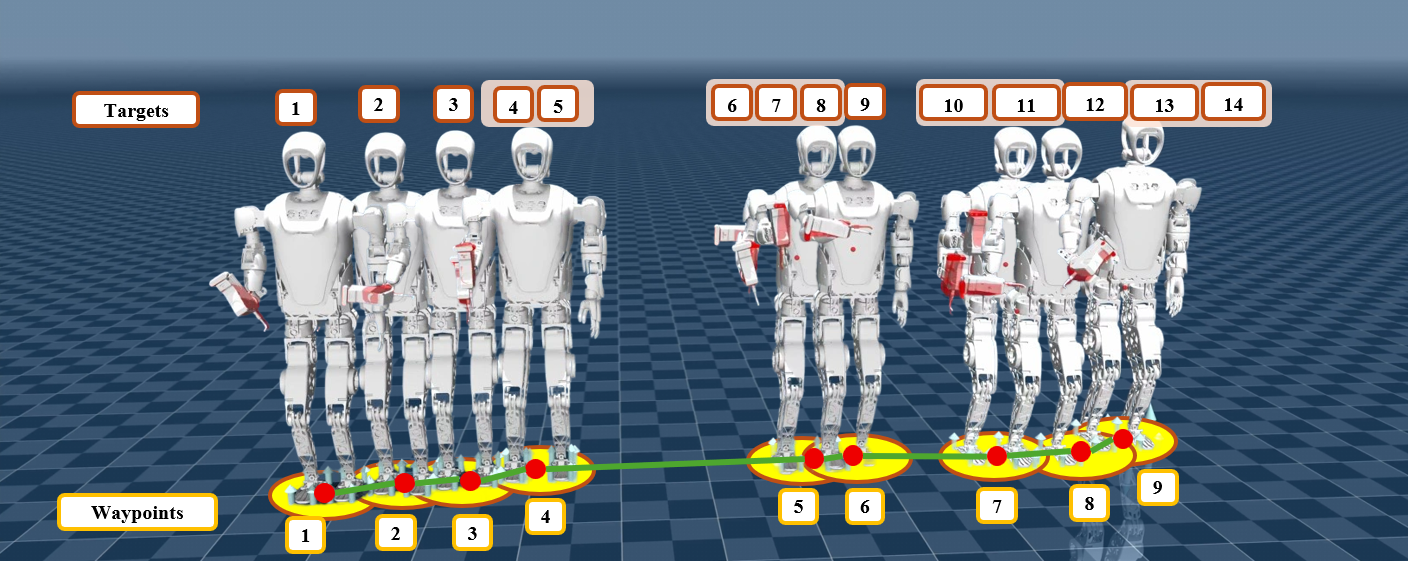}
        \caption{MIP‐based planning: 14 poses scanned in $185s$ with 9 standing positions.}
        \label{fig:sim_exp_mip}
    \end{subfigure}\\
    \begin{subfigure}[b]{\linewidth}
        \includegraphics[width=\linewidth,trim=3cm 3cm 2cm 4cm,clip]{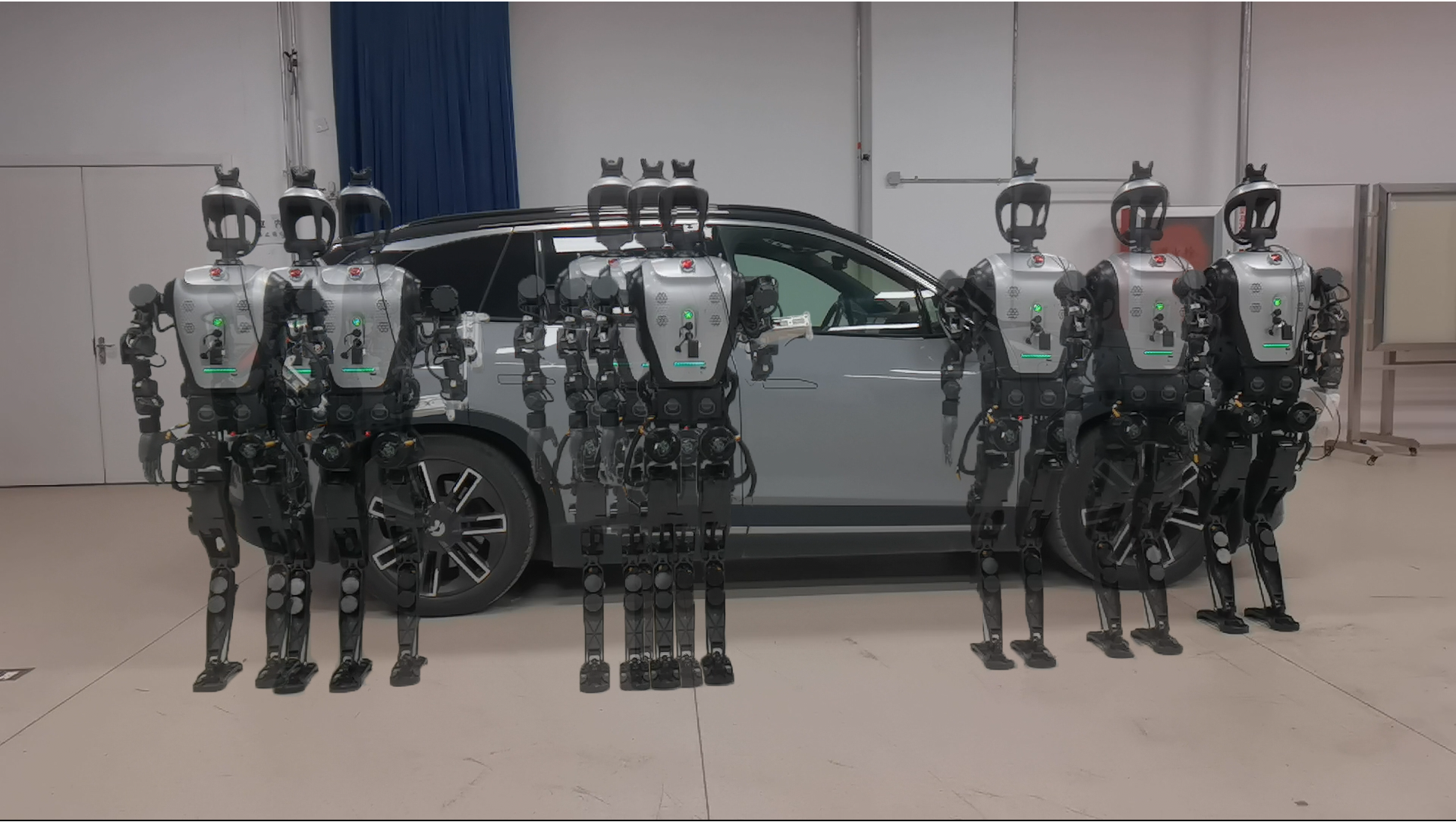}
        \caption{Real-world experiment with proposed MIP-based standing position planning method.}
        \label{fig:phy_exp_mip}
    \end{subfigure}\\
    \caption{\textbf{Comparison of naive and MIP‐based planning for scanning 14 inspection target poses.} Yellow circles denote tolerance regions around each target pose, and the green line indicates the planned robot trajectory.}
    \label{fig:mip_vs_naive}
\end{figure}

In this scenario, the robot was tasked with scanning fourteen inspection target poses sequentially, which were collected from a real-world automotive quality inspection setting. Using our MIP-based planner, these poses were scanned in $185s$ with only nine standing positions, whereas the naive point-to-point method required $265s$ to complete the same task. This represents a 30\% reduction in task completion time. \cref{fig:mip_vs_naive} illustrates the trajectories generated by both methods. The naive point-to-point method plans each target pose independently, leading to 14 standing points and frequent base movements. In contrast, our MIP method clusters reachable targets, reducing standing points to 9. This effectively decreases the number of base relocations, which dominate execution time. By globally optimizing standing points and task sequence, the MIP method significantly improves overall efficiency in dense inspection tasks. 

\subsection{Real-world Validation}

To further validate the practical applicability of our framework, we conducted a real-world experiment involving a vehicle gap inspection scenario. The Kuavo 4Pro robot was tasked with inspecting gaps along the body of a car, requiring it to reach a series of pre-defined inspection points with high precision. We employ an HTC Vive tracker mounted on the robot’s head to provide millimeter-level global localization and use a fixed transformation matrix to convert the head’s 6D pose into the robot’s 2D pose on the plane. The robot successfully completed the inspection task, demonstrating its ability to navigate complex environments and achieve end-effector tracking accuracy with precision. (\cref{fig:phy_exp_mip}) This experiment highlights the potential of our framework for automating inspection tasks in industrial settings.

\section{Discussion \& Conclusion}

This paper introduces a novel and efficient framework for multi-location tasking with humanoid robots in industrial settings. Our key contributions include: a hierarchical planning approach combining IK-based sampling and MIP for reduced complexity; time-optimal standing position generation achieving significant time savings; and an integrated MPC system enabling millimeter-level tracking accuracy. Validated through simulations and physical experiments on the Kuavo 4Pro, our framework demonstrates time efficiency and a high success rate in complex, multi-location tasks. Future work will focus on expanding capabilities through whole-body multi-contact planning to enlarge the workspace and enable loco-manipulation for more dynamic and versatile task execution.
\label{sec:conclusion}

\setstretch{0.97}
{
\small
%\balance
\bibliographystyle{ieeetr}
\bibliography{reference}
}

\end{document}